%% file: plantclef2022.tex
\documentclass[
]{ceurart}

\usepackage{listings}
\begin{document}

\copyrightyear{2022}
\copyrightclause{Copyright for this paper by its authors.
  Use permitted under Creative Commons License Attribution 4.0
  International (CC BY 4.0).}

\conference{CLEF 2022: Conference and Labs of the Evaluation Forum, 
    September 5--8, 2022, Bologna, Italy}


\title{Overview of PlantCLEF 2022: Image-based plant identification at global scale}

\author[1]{Herv\'e Go\"eau}[%
orcid=0000-0003-3296-3795,
email=herve.goeau@cirad.fr
]

\author[1]{Pierre Bonnet}[%
orcid=0000-0002-2828-4389,
email=pierre.bonnet@cirad.fr
]

\author[2]{Alexis Joly}[%
orcid=0000-0002-2161-9940,
email=alexis.joly@inria.fr
]
\address[1]{CIRAD, UMR AMAP, Montpellier, Occitanie, France}
\address[2]{Inria, LIRMM, Univ Montpellier, CNRS, Montpellier, France}

\begin{abstract}
It is estimated that there are more than 300,000 species of vascular plants in the world. Increasing our knowledge of these species is of paramount importance for the development of human civilization (agriculture, construction, pharmacopoeia, etc.), especially in the context of the biodiversity crisis. However, the burden of systematic plant identification by human experts strongly penalizes the aggregation of new data and knowledge. Since then, automatic identification has made considerable progress in recent years as highlighted during all previous editions of PlantCLEF. Deep learning techniques now seem mature enough to address the ultimate but realistic problem of global identification of plant biodiversity in spite of many problems that the data may present (a huge number of classes, very strongly unbalanced classes, partially erroneous identifications, duplications, variable visual quality, diversity of visual contents such as photos or herbarium sheets, etc). The PlantCLEF2022 challenge edition proposes to take a step in this direction by tackling a multi-image (and metadata) classification problem with a very large number of classes (80k plant species). This paper presents the resources and evaluations of the challenge, summarizes the approaches and systems employed by the participating research groups, and provides an analysis of key findings.
\end{abstract}

\begin{keywords}
  LifeCLEF \sep
  fine-grained classification \sep
  species identification \sep
  biodiversity informatics \sep
  evaluation \sep
  benchmark
\end{keywords}
\maketitle

\section{Introduction}

It is estimated that there are more than 300,000 species of vascular plants in the world and new plant species are still discovered and described each year \cite{christenhusz2016number}. This plant diversity has been one of the major elements in the development of human civilization (food, medicine, building materials, recreation, genes, etc.) and it is known to play a crucial role in the functioning and stability of ecosystems \cite{naeem2009biodiversity}. However, our knowledge of plants at the species level is still in its infancy. For the vast majority of species, we have no idea of their specific ecosystemic role or their potential use by humans. Even our knowledge of the geographic distribution and abundance of populations remains very limited for most species \cite{cronk2016plant}. The biodiversity informatics community has made significant efforts over the past two decades to develop global initiatives, digital platforms and tools to help biologists organize, share, visualize and analyze biodiversity data \cite{parr2014encyclopedia, wheeler2004if}. However, the burden of systematic plant identification severely penalizes the aggregation of new data and knowledge at the species level. Botanists, taxonomists and other plant experts spend a lot of time and energy identifying species when their expertise could be more useful in analyzing the data collected. 

As already discussed in \cite{Gaston29042004}, the routine identification of specimens of previously described species has many of the characteristics of other humankind activities that have been automated successfully in the past. Since then, automated identification has made considerable progress, particularly in recent years, thanks to the development of deep learning and Convolutional Neural Networks (CNN) in particular \cite{lecun2015deep}. The long-term evaluation of automated plant identification organized as part of the LifeCLEF initiative \cite{joly2019biodiversity}, illustrated how the arrival of CNNs has impacted performance in a few years. In 2011, the accuracy of the best system evaluated was barely 57\% on a very simple classification task involving 71 species photographed under very homogeneous conditions (scans or photos of leaves on a white background). In 2017, the accuracy of the best CNN was 88.5\% on a much more complex task related to 10K plant species illustrated by highly imbalanced, heterogeneous and noisy visual data \cite{plantclef2017}. In 2018, the best system was able to provide more accurate results than 5 of the 9 specialists who were asked to re-identify a subset of the test images \cite{expertclef2018}.

Thanks to their fast growing audience, existing plant identification applications are an opportunity for high-throughput biodiversity monitoring and for the aggregation of new specific knowledge \cite{nugent2018inaturalist,waldchen2018automated,affouard2017pl}. However, they face the problem of being either too restricted to the flora of particular regions or of being limited to the most common species. As there are more and more species with a transcontinental range (such as naturalized alien species \cite{van2019global} or cultivated plants), fragmenting the identification in regional floras is less and less a reliable approach. On the other hand, focusing only on the most common species on Earth is clearly not a better idea in terms of biodiversity. 

The PlantCLEF2022 challenge edition proposes to take a step in this direction by tackling a multi-image (and metadata) classification problem with a very large number of classes (80k plant species). CNNs and the recent Vision Transformers techniques are without doubt the most promising solution today for addressing such a very large scale image classification task. However, no previous work had reported image classification results of this order of magnitude, whether or not they are biological entities. This paper presents the resources and evaluations of the challenge, summarizes the approaches and systems employed by the participating research groups, and provides an analysis of key findings.

\section{Dataset}

To evaluate the above mentioned scenario at a large scale and in realistic conditions, we built and shared two training datasets: "trusted" and "web" (i.e. with or without species labels provided and checked by human experts), totaling 4M images on 80k classes coming from different sources.\\
\\
\textbf{Training set "trusted"}: this training dataset is based on a selection of more than 2.9M images covering 80k plant species shared and collected mainly by GBIF (and EOL to a lesser extent). These images come mainly from academic sources (museums, universities, national institutions) and collaborative platforms such as inaturalist or Pl@ntNet, implying a fairly high certainty of determination quality. Nowadays, many more photographs are available on these platforms for a few thousand species, but the number of images has been globally limited to around 100 images per species, favouring types of views adapted to the identification of plants (close-ups of flowers, fruits, leaves, trunks, ...), in order to not unbalance the classes and to not explode the size of the training dataset.\\
\\
\textbf{Training set "web":} in contrast, the second data set is based on a collection of web images provided by search engines Google and Bing. This initial collection of several million images suffers however from a significant rate of species identification errors and a massive presence of duplicates and images less adapted for visual identification of plants (herbariums, landscapes, microscopic views...), or even off-topic (portrait photos of botanists, maps, graphs, other kingdoms of the living, manufactured objects, ...). The initial collection has been then semi-automatically revised to drastically reduce the number of these irrelevant pictures and to maximise, as for the trusted dataset, close-ups of flowers, fruits, leaves, trunks, etc. The "web" dataset finally contains about 1.1 million images covering around 57k species.\\
\\
\textbf{Test set:} is built from multi-image plant observations collected on the Pl@ntNet platform during the year 2021 (observations not yet shared through GBIF, and thus not present in the training set). Only observations that received a very high confidence score in the Pl@ntNet collaborative review process were selected for the challenge to ensure the highest possible quality of determination. This process involves people with a wide range of skills (from beginners to world-leading experts), but these have different weights in the decision algorithms. Finally, the test set contains about 27k plant observations related to about 55k images (a plant can be associated with several images) covering about 7.3k species.\\
\\
Table\ref{tab:datastats} shows various statistics about the three datasets. We can note a significant difference between the number of species present in the training sets and the test set mainly due to the fact that it was difficult to collect so much expert data by botanists at such a scale. However, having fewer species in the test set remains consistent with a realistic scenario faced by automatic identification systems such as Pl@ntNet, Inaturalist: these systems must be able to recognize as many species as possible without knowing in advance which species will be the most frequently requested and which will never be requested.

\input{tables/stats}

\section{Task Description}

The challenge was hosted in the AICrowd plateform\footnote{\url{https://www.aicrowd.com/challenges/lifeclef-2022-plant}}. The task was evaluated as a plant species retrieval task based on multi-image plant observations from the test set. The goal was to retrieve the correct plant species among the top results of a ranked list of species returned by the evaluated system. The participants had access to the training set in mid-February 2022, the test set was published 6 weeks later in early April, and the round of submissions was then open during 5 weeks.

The metric used for the evaluation of the task is the Macro Average (by species) Mean Reciprocal Rank (MA-MRR). The Mean Reciprocal Rank (MRR) is a statistic measure for evaluating any process that produces a list of possible responses to a sample of queries ordered by probability of correctness. The reciprocal rank of a query response is the multiplicative inverse of the rank of the first correct answer. The MRR is the average of the reciprocal ranks for the whole test set:
\begin{equation}
    MRR = \frac{1}{O} \sum_{i=1}^O \frac{1}{\text{rank}_i}
\end{equation} 
where $O$ is the total number of plant observations (query occurrences) in the test set and $\text{rank}_i$ is the rank of the correct species of the plant observation $i$.\\

However, the Macro-Average version of the MRR (average MRR per species in the test set) was used because of the long tail of the data distribution to rebalance the results between under- and over-represented species in the test set:
\begin{equation}
    MA-MRR = \frac{1}{S} \sum_{j=1}^S \frac{1}{O_j} \sum_{i=1}^{O_j} \frac{1}{\text{rank}_i}
\end{equation} 
where $S$ is the total number of species in the test set, $O_j$ is the number of plant observations related to a species $j$.

\section{Participants and methods}

90 research groups registered to the LifeCLEF plant challenge 2022. Among this large raw audience, 8 research groups finally succeeded in submitting run files. Details of the used methods and evaluated systems are synthesized below and further developed in the working notes of the participants (Mingle Xu \cite{MingleXu2022}, Neuon AI\cite{NeuonAI2022}, Chans Temple \cite{ChansTemple12022}, BioMachina \cite{BioMachina2022}, KL-SSN-CE \cite{KLSSNCE2022} and SVJ-SSN-CE \cite{SVJSSNCE2022}). Table \ref{tab:rawresults} reports the results obtained by each run as well as a brief synthesis of the methods used in each of them. Complementary, the following paragraphs give a few more details about the methods and the overall strategy employed by each participant (the paragraphs are sorted in descending order of the best score obtained by each team; the number of runs does not reflect the total number of submitted runs but the ones described in the participants' working notes).\\
\\
\textbf{Mingle Xu, South Korea, 7 runs, \cite{MingleXu2022}}: this team based their work on a vision transformer pre-trained with a Self Supervised Learning (SSL) method, a recent and increasingly popular approach in the field of computer vision. This type of approach is quite disruptive since it does not use labels compared to the usual Supervised Transfer Learning (STL) method where typically a network is pre-trained first to perform a classification task on a generic dataset such as ImageNet, and then finetuned on a specific dataset. It is expected that a network pre-trained with an SSL method extract better features, with more generalization power, which can then be then efficiently finetuned in a supervised manner on various downstream tasks such as image classification or object detection. SSL methods generally work with two models (ViT or CNN), for instance with a "student" model that tries to extract similar features learned by a "teacher" model despite several alterations of the image (data augmentation) such as for DINO \cite{caron2021emerging}. The Masked Auto-Encoder (MAE)  \cite{he2022masked} used by the team is an other way to perform a self-supervised learning inspired by the successful idea of masked language modeling in Natural Language Processing, especially since BERT \cite{devlin2018bert}. The masking process was difficult to apply to CNN-based architectures whereas it becomes quite straightforward with vision transformers since they work internally in the form of visual patches or "tokens" with positional embedding. MAE is similar to BEIT \cite{bao2021beit} where the self supervised task consists in training a backbone vision transformer to predict missing tokens from partially masked images. Beyond the originality of the pre-training method, the successive runs of this team consist in following snapshots over several days of training.\\
\\
\textbf{Neuon AI, Malaysia, 4 runs, \cite{NeuonAI2022}}: this participant used various ensembles of models finetuned on the "trusted", and from some of them on also the "web", training dataset and based on Inception-v4 and Inception-ResNet-v2 architectures \cite{szegedy2016inception}. Most of the models are directly finetuned CNNs but as a multi-task classification network related to the five taxonomy levels: Species (the main task), Genus, Family, Order and Class (in the botanical sense). A second type of model is triplet network based also on a Inception-v4 or Inception-ResNet-v2 CNN models where the last fully connected layer is used for embedding representation limited to 500 visual words instead of the heavy 80k outputs typically necessary for the species classification task. However, the triplet network seems to be longer to saturate the training and worked only on the species levels making difficult to compare the real contribution of this type of network in the ensemble results submitted by this team.\\
\\
\textbf{Chans Temple, Malaysia, 3 runs, \cite{ChansTemple12022}}: inspired by the Hierarchical Deep Convolutional Neural Networks (HD-CNN) in \cite{yan2015hd}, this participant explored the taxonomy information in different ways with a more recent architecture and deep learning framework. A first strategy is to first finetune a ResNet34 model on a family classification task and then finetune it to the species level. A second similar strategy begins with a multi-task classification by finetuning a ResNet34 model on all five levels of taxonomy at once (species, genus, family, order, botanical class) before continuing with finetuning on the species level only. The best submission was an ensemble of these two models combined with two regular models without taxonomy (a ResNet50, and a ResNet50-Wide).\\
\\
\textbf{BioMachina, Costa Rica, 5 runs, \cite{BioMachina2022}}: the main contribution and very interesting idea explored by this team was to let a model learn its own hierarchy instead exploiting directly the taxonomy provided in the dataset. They proposed a 2-level hierarchical softmax which has the interesting property to reduce drastically the weights of the usual fully connected layer of the classification head while maintaining the same performances. This property is illustrated by comparing a EfficientNetB4 with its learnt hierarchical version. Considering that the output size of a EfficientNetB4 backbone is 2048, the proposed hierarchical design allows to reduce from 160 millions of parameters in a fully connected layer (2048x80k weights + 80k biases) to 28.2 millions of parameters, enabling potentially a much more faster training since it allows to increase the batch size. Aside this hierarchical approach, they also highlighted on a standard ResNet50 that a good strategy of training was first to finetune a model on the web training dataset before finetuning it on the trusted dataset. The results obtained with a heavier ResNet101 model on the other hand did not reveal to have a great impact. Following the modern best practices of deep learning, various techniques to reduce learning times and ensure better convergence of models during learning have also been explored (automatic mixed precision, batch accumulation, gradient clipping).\\
\\
\textbf{KL-SSN-CE, India, 1 runs, \cite{KLSSNCE2022}}: this team tried to train in a recent deep learning framework the famous historical model AlexNet which marked the revival of neural networks 10 years ago. They compared many optimizers and loss functions, and selected for their main submission AdaGrad and KL Divergence.\\
\\
\textbf{SVJ-SSN-CE, India, 2 runs, \cite{SVJSSNCE2022}}: this team focused their contribution on the classification head on the top of a ResNet50 pre-trained model. After disappointing results with a standard classifier using only one Fully Connected (FC) layer, they expected more relevant features and better classification performances by adding a second intermediate FC layer and using a sparse categorical cross-entropy loss. In a second multi-level classification ResNet50 model, they implemented a probabilistic tree approach to use the taxonomy information.

\section{Results}

\input{tables/results_table}

\begin{figure}[t]
\centering
\includegraphics[width=0.95\linewidth]{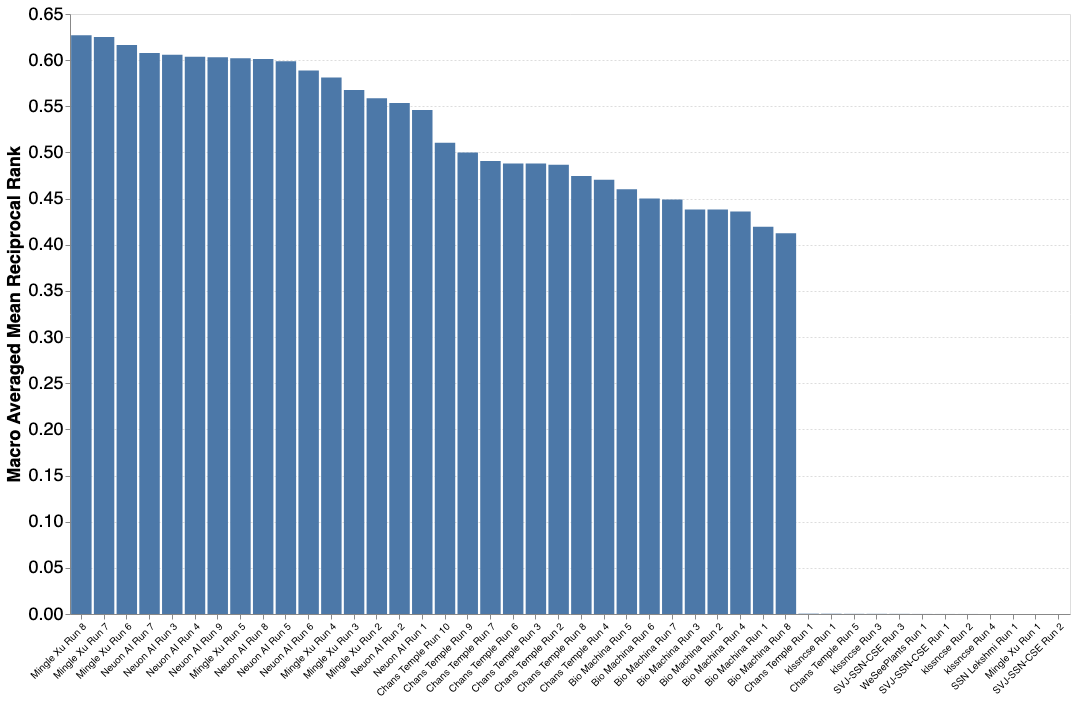}
\caption{Scores achieved by all systems evaluated within the plant identification task of LifeCLEF 2022}
\label{fig:PlantCLEF2022OfficialScore}
\end{figure}

We report in Figure \ref{fig:PlantCLEF2022OfficialScore} the performance achieved by the collected runs. Table \ref{tab:rawresults} provides the results achieved by each run as well as a brief synthesis of the methods used in each of them. The main outcomes we can derive from that results are the following:
\\
\\
\textbf{A new supremacy of the vision transformer:}  the best results were obtained by the only team which used vision transformers \cite{MingleXu2022} contrary to the others which used convolutional neural networks, i.e. the traditional approach of the state-of-the-art for image-based plant identification. If we compare the performance of the best vision transformer (Mingle XU Run 8, MA-MRR=0.626) to the one of the best CNN trained on the same data (Neuon AI Run 2, MA-MRR=0.553), we can observe that the gain is very high.\\ 
\\
\textbf{The race for GPUs}: however, the gain in identification performance obtained by the vision transformers is paid for by a significant increase of the training time. The winning team reported that they had to stop the training of the model in order to submit their run to the challenge. Thus, better results could have surely been obtained with a few more days of training (as demonstrated through post-challenge evaluations reported in the their working note \cite{MingleXu2022}. 
\\
\\
\textbf{Rationalization of the use of memory:} One of the main difficulties of the challenge was the very large number of classes (80K). For most of the models used, the majority of the weights to be trained are those of the last fully connected layer of the classifier. This was an important consideration for all participants in their model selection strategy. Some teams have tried to limit this cost through specific approaches. The BioMachina team \cite{BioMachina2022}, in particular, used a two-level hierarchical softmax to reduce the number of weights drastically. They reported an considerable training time reduction while maintaining almost the same identification quality.\\
\\
\textbf{Taxonomy can help}: 3 teams worked with the taxonomy, with multi-task classification network (Neuon AI, Chans Temple) or for manipulating probabilities (KL-SSN-CE) or for pre-training a model (Chans Temple, BioMachina). Chans Temple highlighted in his preliminary results on a validation set that pretraining a model with the Family level, and in a lesser extent all the taxonomy levels, improve the performance of a single species classification task. But all these approach are at the expense of additional layers or/and computing time, contrary to the original BioMachina's approach which aims at reducing the memory footprint by letting a model to learn its own hierarchy.\\
\\
\textbf{The noisy web training dataset may help}: as yet noticed in PlantCLEF 2017 \cite{plantclef2017} Neuon AI showed again that the noisy data from the web training dataset does improve the generalisation of their CNN models, as well as BioMachina who successfully pre-training models on the web training dataset before finetuning them on the trusted dataset.

\section{Additional analyses}

\begin{figure}[t]
\centering
\includegraphics[width=\linewidth]{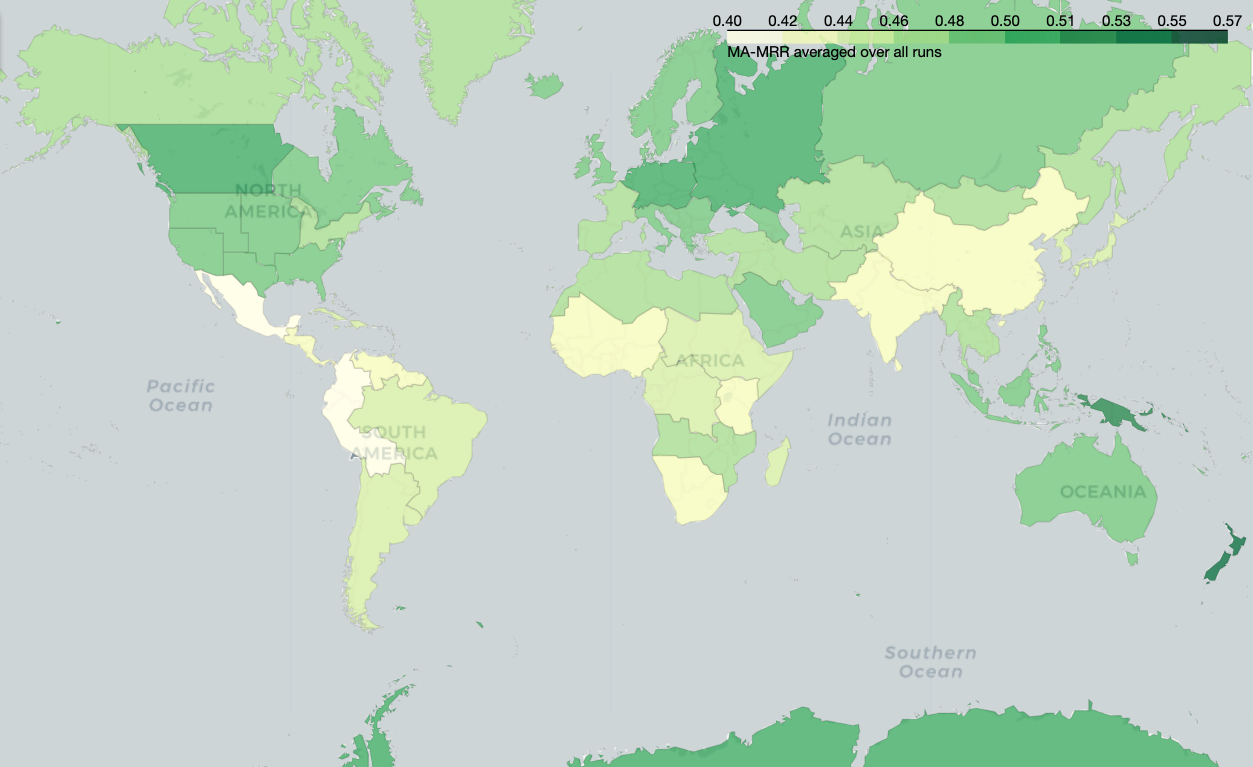}

\caption{Map of MA-MRR averaged over all runs by regional species checklists (World Geographical Scheme for Recording Plant Distributions - WGSRPD - TDWG level 3)}
\label{fig:checklist_results_map}
\end{figure}

During the previous years of PlantCLEF, it was shown that it was much more difficult to identify species from the Amazon rainforest than species from Europe and North America \cite{plantclef2019}. By extension, we can assume that most identification systems would encounter the same difficulties in other tropical rainforests in Equatorial Africa or Indonesia, but without really measuring this. This year's global challenge can give us some information and a first overview of the performance of automatic systems in different large regional areas. Table \ref{tab:checklistresults2} and Figure \ref{fig:checklist_results_map} show the MA-MRR averaged over all runs submitted and detailed for different regional species checklists. The regional division follows the level 3 of the standard WGSRPD (World Geographical Scheme for Recording Plant Distributions \cite{brummitt2001world} managed by the 
Biodiversity Information Standards (TDWG).

In this table, we can see that globally the areas corresponding to the western countries (Europe, North America, Australia and New Zealand) obtain performances among the highest, while in the lower part of the table we can note many areas corresponding to tropical regions from South and Central America, India and Africa. This result is to be expected since the average number of images per species tends to be correlated with the average run performances. However, it is harder to explain some good performances like those obtained over Papuasia, a typically tropical region with less data. This type of analysis would deserve to be more developed and detailed in order to draw useful conclusions that could be used in the future as recommendations for new data collection efforts around the world.

\input{tables/checklist_results2}

\section{Conclusion}

This paper presented the overview and the results of the LifeCLEF 2022 plant identification challenge following the 11 previous ones conducted within CLEF evaluation forum. This year the task was performed on the biggest plant images dataset ever published in the literature. This dataset was composed of two distinct sources: a trusted set built from the GBIF and a noisy web dataset totaling both 4M images and covering 80k species. The main conclusion of our evaluation is that vision transformers performed better than CNNs as demonstrated by the Mingle Xu team knowing that the training of their model was not yet completed at the time of the challenge closure. This shows the potential of these techniques on huge datasets such as the one of PlantCLEF. However, training those models requires more computational resources that only participants with access to large computational clusters can afford. Beyond the duality between free vision transformers and CNNs, the BioMachina team has demonstrated that it is possible to drastically reduce the number of parameters of the classification head at the output of a backbone while maintaining the performance of a classical approach with a fully connected layer. This result is of great importance because it allows to consider more serenely in the future classification models of species that would address the 300,000 species of plants on a global scale. This contribution is also interesting because it allows us to redefine an optimized hierarchy for visual feature extraction mechanisms by moving away from the expert hierarchical framework of taxonomy, which may be a bit too rigid because it does not systematically reflect morphological similarities between species but also functional trait proximities and phylogenetic relationships. Beyond the technical aspects, the complementary analysis of the performances detailed by sub-continental the world would deserve to be more developed and detailed in order to draw useful conclusions of high importance in botany and biodiversity informatics in general, such as future recommendations for new data collection efforts around the world.

\end{document}

%% file: tables/stats.tex
\begin{table}[!h]
    \caption{Statistics of the LifeCLEF 2022 Plant Identification Task: "n/s" means not specified }
    \centering
    \begin{tabular}{|r|r|r|r|r|r|r|}
    \hline
    Dataset         & Images    & Observations  & Classes (species) & Genera    & Families  & Orders \\
    \hline
    Train "trusted"	& 2,886,761 & n/s           & 80,000	        & 9,603     & 483       & 84 \\
    Train "web"     & 1,071,627 & n/s           & 57,314            & 8,649     & 479       & 84 \\
    Test            & 55,307    & 26,869        & 7,339             & 2,527     &           &  \\
    \hline
\end{tabular}
\label{tab:datastats}
\end{table}

%% file: tables/results_table.tex
\begin{table}[h]
    \caption{Results of the LifeCLEF 2022 Plant Identification Task (limited to the runs described in the participants' working notes). \textbf{Architecture:} AN: AlexNet, EN4: EfficientNetB4, HEN4: Hierarchical EfficientNetB4, IRv2: Inception-ResNet-v2, Iv4: Inception-v4, RN: ResNet, Tr: triplet network, ViT-L: Vision Transformer Large. \textbf{Datasets:} IN1k: ImageNet1k, PlantCLEF2022: T (Trusted), W (Web), TW (Trusted \& Web). \textbf{Pre-training methods:} SSL: Self Supervised Learning, STL: Supervised Transfer Learning. \textbf{Taxonomy:} - (no), A (All: species, genus, family, order, class), F: family, LH: Learned Hierarchy)}
    \centering
    \begin{tabular}{|l|c|c|c|c|c|}
    \hline
    Team run name	    & Architecture  & Pre-training  & Training & Taxonomy & MA-MRR\\
    \hline
    Mingle Xu Run 8     & ViT-L         & SSL MAE IN1k  & T     & -         & 0.62692\\
    Mingle Xu Run 7     & ViT-L         & SSL MAE IN1k	& T     & -         & 0.62497\\
    Mingle Xu Run 6	    & ViT-L	        & SSL MAE IN1k	& T     & -         & 0.61632\\
    Neuon AI Run 7	    & Iv4, IRv2	    & STL IN1k	    & TW    & -         & 0.60781\\ 
    Neuon AI Run 3	    & Iv4, IRv2	    & STL IN1k	    & TW    & A         & 0.60583\\
    Neuon AI Run 4	    & Iv4, IRv2	    & STL IN1k      & TW    & A         & 0.60381\\
    Neuon AI Run 9	    & Iv4, IRv2	    & STL IN1k	    & TW    & A         & 0.60301\\
    Mingle Xu Run 5	    & ViT           & SSL MAE IN1k	& T	    & -         & 0.60219\\
    Neuon AI Run 8	    & Iv4, IRv2	    & STL IN1k	    & TW    & A         & 0.60113\\
    Neuon AI Run 5	    & Iv4, IRv2, Tr & STL IN1k	    & TW    & A         & 0.59892\\
    Neuon AI Run 6	    & IRv2	        & STL IN1k  	& TW    & A         & 0.58874\\
    Mingle Xu Run 4	    & ViT-L	        & SSL MAE IN1k	& T     & -         & 0.58110\\
    Mingle Xu Run 3	    & ViT-L	        & SSL MAE IN1k	& T     & -         & 0.56772\\
    Mingle Xu Run 2	    & ViT-L	        & SSL MAE IN1k	& T     & -         & 0.55865\\
    Neuon AI Run 2	    & Iv4, IRv2	    & STL IN1k	    & T     & A         & 0.55358\\
    Neuon AI Run 1	    & IRv2	        & STL IN1k	    & T     & A         & 0.54613\\
    Chans Temple Run 10	& RN34, RN50	& STL IN1k		& T     & -,A,F     & 0.51043\\
    Chans Temple Run 9	& RN34	        & STL IN1k      & T     & F	        & 0.49994\\
    Chans Temple Run 8	& RN34	        & STL IN1k		& T	    & A         & 0.47447\\
    BioMachina Run 5	& RN50  & STL IN1k$\rightarrow$W& T     & -         & 0.46010\\
    BioMachina Run 6	& RN101 & STL IN1k$\rightarrow$W& T     & -         & 0.45011\\
    BioMachina Run 3	& RN50	        & STL IN1k	    & T     & -         & 0.43820\\
    BioMachina Run 1	& HEN4          & STL IN1k      & T     & LH        & 0.41950\\
    BioMachina Run 8	& EN4           & STL IN1k      & T     & -         & 0.41240\\
    KL-SSN-CE Run 1     & AN 	        & STL IN1k	    & T     & -         & 0.00029\\
    SVJ-SSN-CSE Run 3	& RN50	        & STL IN1k	    & T     & A         & 0.00015\\
    SVJ-SSN-CSE Run 1	& RN50	        & STL IN1k	    & T     & A         & 0.00005\\
\hline
\end{tabular}
\label{tab:rawresults}
\end{table}

%% file: tables/checklist_results2.tex
\begin{table}[!h]
    \caption{MA-MRR averaged over all runs by regional species checklists (World Geographical Scheme for Recording Plant Distributions - WGSRPD - TDWG level 3). Img/sp gives the average number of trusted images per species on the current checklist.}
    \scriptsize
    \centering
    \begin{tabular}{|r|c|c|c|}
    \hline
    Checklist	&	Species	& Img/sp		&	Mean MA-MRR\\
    \hline
    Middle Atlantic Ocean	&	299	&	96	&	0.5706\\
    New Zealand	&	809	&	97	&	0.5411\\
    South-Central Pacific	&	479	&	94	&	0.5351\\
    Papuasia	&	439	&	89	&	0.5204\\
    Northwestern Pacific	&	210	&	93	&	0.5150\\
    North-Central Pacific	&	448	&	95	&	0.5094\\
    Western Canada	&	739	&	99	&	0.5038\\
    Subantarctic Islands	&	268	&	98	&	0.5029\\
    Eastern Europe	&	1916	&	97	&	0.5029\\
    Middle Europe	&	2537	&	94	&	0.4988\\
    Antarctic Continent	&	2	&	68	&	0.4974\\
    Eastern Canada	&	934	&	98	&	0.4951\\
    Caucasus	&	1265	&	96	&	0.4949\\
    Northern Europe	&	1876	&	97	&	0.4946\\
    Southeastern Europe	&	3055	&	91	&	0.4942\\
    South-Central U.S.A.	&	1032	&	98	&	0.4893\\
    Australia	&	1098	&	93	&	0.4878\\
    Northwestern U.S.A.	&	1041	&	98	&	0.4864\\
    Southwestern Pacific	&	619	&	92	&	0.4861\\
    Southeastern U.S.A.	&	1766	&	96	&	0.4835\\
    Malesia	&	880	&	86	&	0.4831\\
    Southwestern U.S.A.	&	1176	&	97	&	0.4812\\
    North-Central U.S.A.	&	1290	&	98	&	0.4807\\
    Arabian Peninsula	&	701	&	88	&	0.4807\\
    Siberia	&	881	&	98	&	0.4803\\
    Russian Far East	&	746	&	96	&	0.4765\\
    Northeastern U.S.A.	&	1507	&	98	&	0.4760\\
    Indo-China	&	1142	&	86	&	0.4750\\
    Macaronesia	&	1344	&	94	&	0.4737\\
    Southwestern Europe	&	3201	&	90	&	0.4736\\
    South Tropical Africa	&	875	&	84	&	0.4728\\
    Subarctic America	&	440	&	99	&	0.4722\\
    Western Asia	&	2093	&	93	&	0.4718\\
    Northern Africa	&	1846	&	91	&	0.4665\\
    Middle Asia	&	1202	&	96	&	0.4659\\
    Mongolia	&	352	&	97	&	0.4605\\
    Southern South America	&	1575	&	82	&	0.4538\\
    West-Central Tropical Africa	&	908	&	82	&	0.4531\\
    Brazil	&	1142	&	79	&	0.4523\\
    Northeast Tropical Africa	&	1016	&	81	&	0.4513\\
    Eastern Asia	&	1376	&	91	&	0.4503\\
    Western Indian Ocean	&	929	&	84	&	0.4461\\
    Caribbean	&	1418	&	88	&	0.4404\\
    Northern South America	&	921	&	84	&	0.4379\\
    Indian Subcontinent	&	1661	&	89	&	0.4363\\
    West Tropical Africa	&	774	&	82	&	0.4332\\
    Central America	&	1215	&	87	&	0.4329\\
    China	&	1475	&	86	&	0.4325\\
    Southern Africa	&	1226	&	86	&	0.4311\\
    East Tropical Africa	&	898	&	78	&	0.4291\\
    Mexico	&	1477	&	88	&	0.4071\\
    Western South America	&	1655	&	80	&	0.4021\\
    \hline
\end{tabular}
\label{tab:checklistresults2}
\end{table}